\documentclass{article}
\usepackage[nonatbib,final]{neurips_2020}
\usepackage{times}
\usepackage{url}
\usepackage{latexsym}
\usepackage{tikz}
\usetikzlibrary{decorations.pathreplacing}
\usepackage{fancyvrb}
\usepackage{graphicx}
\usepackage{xcolor}
\usepackage{amsmath}
\usepackage{amssymb}
\usepackage{bbm}
\usepackage{multicol}
\usepackage{subfig}
\usepackage{multirow}
\usepackage{footnote}
\usepackage{ifluatex}
\ifluatex
\usepackage{pdftexcmds}
\makeatletter
\let\pdfstrcmp\pdf@strcmp
\let\pdffilemoddate\pdf@filemoddate
\makeatother
\fi
\usepackage{svg}
\usepackage{pdfpages}
\newcommand{\attr}[1]{{\tt \small #1}}
\makeatletter
\newcommand\footnoteref[1]{\protected@xdef\@thefnmark{\ref{#1}}\@footnotemark}
\makeatother

\newcommand{\cev}[1]{\reflectbox{\ensuremath{\vec{\reflectbox{\ensuremath{#1}}}}}}

\title{Is the User Enjoying the Conversation? A Case Study on the Impact on the Reward Function.}
\author{Lina M. Rojas-Barahona. \\
  Orange-Labs. Lannion\\
  2 Avenue Pierre Marzin\\
  22300, Lannion France. \\
  \texttt{linamaria.rojasbarahona@orange.com}} 
\begin{document}

\maketitle
\begin{abstract}
The impact of user satisfaction in policy learning task-oriented dialogue systems has long been a subject of research interest. 
Most current models for estimating the user satisfaction either (i) treat out-of-context short-texts, such as product reviews, or (ii) rely on turn features instead of on distributed semantic representations. 
In this work we adopt deep neural networks that use distributed semantic representation learning for estimating the user satisfaction in conversations.  We evaluate the impact of modelling context length in these networks. Moreover, we show that the proposed hierarchical network outperforms state-of-the-art quality estimators. Furthermore, we show that applying these networks to infer the reward function in a Partial Observable Markov Decision Process (POMDP) yields to a great improvement in the \textit{task success rate}. %when applied to the same domain on which the estimators were trained. %Hierarchical networks are also competitive enough to initialise unknown\footnote{\label{f:dom} A domain is \textit{known} if it is the domain on which the satisfaction estimators were trained, all the other domains are \textit{unknown}.} domains that can be fine-tuned later on.
\end{abstract}
%~\cite{walker1997paradise,schmitt2011modeling}
%rojas2016deep

\section{Introduction}
\label{intro}
The impact of user satisfaction in policy learning for task-oriented dialogue systems has long been a subject of research interest~\cite{walker1997paradise,schmitt2011modeling,ultes-etal-2015-quality,ultes-2019-improving}.
Similarly, sentiment analysis has been widely adopted to analyse massive blogs, recommendations tweets and reviews~\cite{rojas2016deep,do2019deep}. Although sentiment analysis can be used to infer user satisfaction, most of the work in the literature that incorporates sentiment analysis in dialogue has focused on the generation of empathetic responses in end-to-end models for chitchat~\cite{lee2018scalable,ma2020survey}. Moreover, most sentiment analysis solutions focus on the analysis of out-of-context short texts~\cite{rojas2016deep}.  In this work we are interested in the study of user satisfaction for measuring the quality of the interaction in task-oriented dialogues, in which dialogue is modelled as a POMDP~\cite{young2013pomdp}. 
    
Our contribution in this paper is twofold. First we study distinct deep learning networks that use distributed representations for predicting satisfaction scores. At this stage we would like to answer the following question: does relying only on distributed representations improve the performance of neural models? Therefore we evaluate the performance of hierarchical networks and state-of-the-art Transformers initialised with pre-trained word and contextual representations respectively.

Second, we evaluate the impact of using the best trained network for computing the reward function within a POMDP dialogue framework~\cite{ultes2017pydial}.
We would like to determine how realistic it is to incorporate user satisfaction estimators that rely solely on distributional semantics in reinforcement learning (RL) dialogue systems. This approach can be used for instance to train satisfaction predictors from large human-human chats, in which satisfaction has been self-scored by users.
%that can be integrated later on in a reinforcement learning (RL) dialogue framework for computing the reward function.
%, on the Let's Go domain~\cite{raux2005let},

Our case study is the English LEGO corpus of human-machine spoken conversations~\cite{schmitt-etal-2012-parameterized}, which has been annotated at each system turn with the interaction quality (IQ), ranging from 1 (poor quality) to 5 (good quality). Our results suggest that distributed representations do outperform state-of-the-art models trained on fine-tuned features. We also show that using IQ estimators in the reward function greatly improves the task success rate for dialogues in the same domain the networks were trained on, which in this case is the Let's Go domain~\cite{raux2005let}. %Moreover, the task success rate is competitive enough to suggest that these estimators generalise well and that they can be used to initialise and fine-tuned user-satisfaction models on unknown domains\footnoteref{f:dom}.

In the remainder of this paper we present the related work in Section~\ref{back}. We introduce the user satisfaction estimators in Section~\ref{estimators}. The Reward function and POMDP dialogue framework is explained in Section~\ref{pomdp}. The results of our experiments are described in Section~\ref{exps}. Finally the discussion and conclusions are presented in Section~\ref{disc}.
%
% The following footnote without marker is needed for the camera-ready
% version of the paper.
% Comment out the instructions (first text) and uncomment the 8 lines
% under "final paper" for your variant of English.
% 
%\section{}

\section{Related Work}
\label{back}

The prediction of the IQ score for the LEGO corpus was first presented in~\cite{schmitt2015interaction} with a Support Vector Machine (SVM) baseline. More recently a Bi-directional Long-Short Term Memory (BiLSTM) network to predict the IQ was proposed in~\cite{ultes-2019-improving}. Both the SVM and the BiLSTM were trained on fine-tuned \textit{turn features} such as \attr{the automatic Speech Recognition (ASR) success}, \attr{the ASR mean confidence}, \attr{type of system action} and \attr{the number of rejections}. The IQ predictors presented in~\cite{ultes-2019-improving} were applied to task-oriented dialogue domains other than the Let's Go domain (e.g. Cambridge restaurants, Cambridge hotels and San Francisco restaurants) for computing the reward within a POMDP policy learning framework.  

Unlike~\cite{ultes-2019-improving,schmitt2015interaction} the models proposed in this paper rely solely on \textit{distributed semantic representations} instead of on \textit{turn features}. Turn features are not always available when using proprietary ASR services, which typically merely return the best transcription. In addition ASR features are not applicable to chatbots.
Inspired by~\cite{ultes-2019-improving} we are also interested in evaluating the impact of quality estimators on the reward function for dialogue policy learning but applied to the Let's Go domain with simple and complex constraints (Table~\ref{t:domains}).
%but not only by affecting the final reward function but also the reward at each dialogue turn. %We evaluate the impact for the Let's Go domain by using PyDial, a multi-domain POMDP dialogue framework~\cite{ultes2017pydial}.

In this paper we evaluate the adoption of Transformers for the estimation of the user satisfaction on dialogue data. Although \cite{henderson2019convert} proposed contextual representations for dialogue, their distributed representations are specialised on retrieval-based approaches, in which the existence of a set of candidate responses at each turn is assumed. Since we aim to apply Transformers to \textbf{task-oriented dialogues} of great length we did not adopt~\cite{henderson2019convert} embeddings in this work. For the same reason, our work is different of generative end-to-end approaches to dialogue~\cite{lee2018scalable,ma2020survey}, since our goal is to build dialogues that successfully achieve a task that has been well defined in a knowledge-base. In this work, generation is only a component of a complex dialogue architecture, not the whole dialogue system (Figure~\ref{achictecture}).

\section{Networks for Estimating the User Satisfaction}
\label{estimators}
We study three distinct neural networks:  hierarchical Gated Recurrent Units(GRUs)~\cite{cho-etal-2014-learning} with attention , Transformers for generating contextual embeddings that feed a GRU layer and solely Transformers~\cite{vaswani2017attention}. We are interested in studying the impact of context-length in transformers, because real dialogues can easily attain a context of thousands of tokens (Table~\ref{t:corp}). Therefore, we explore BERT~\cite{devlin2018bert}, DistilBERT~\cite{sanh2019distilbert} and Transformers eXtra-Large (Transformers-XL)~\cite{dai2019transformer}.

\begin{figure}[ht]
\centering
\subfloat[Hierarchical GRUs]{\includegraphics[scale=0.4]{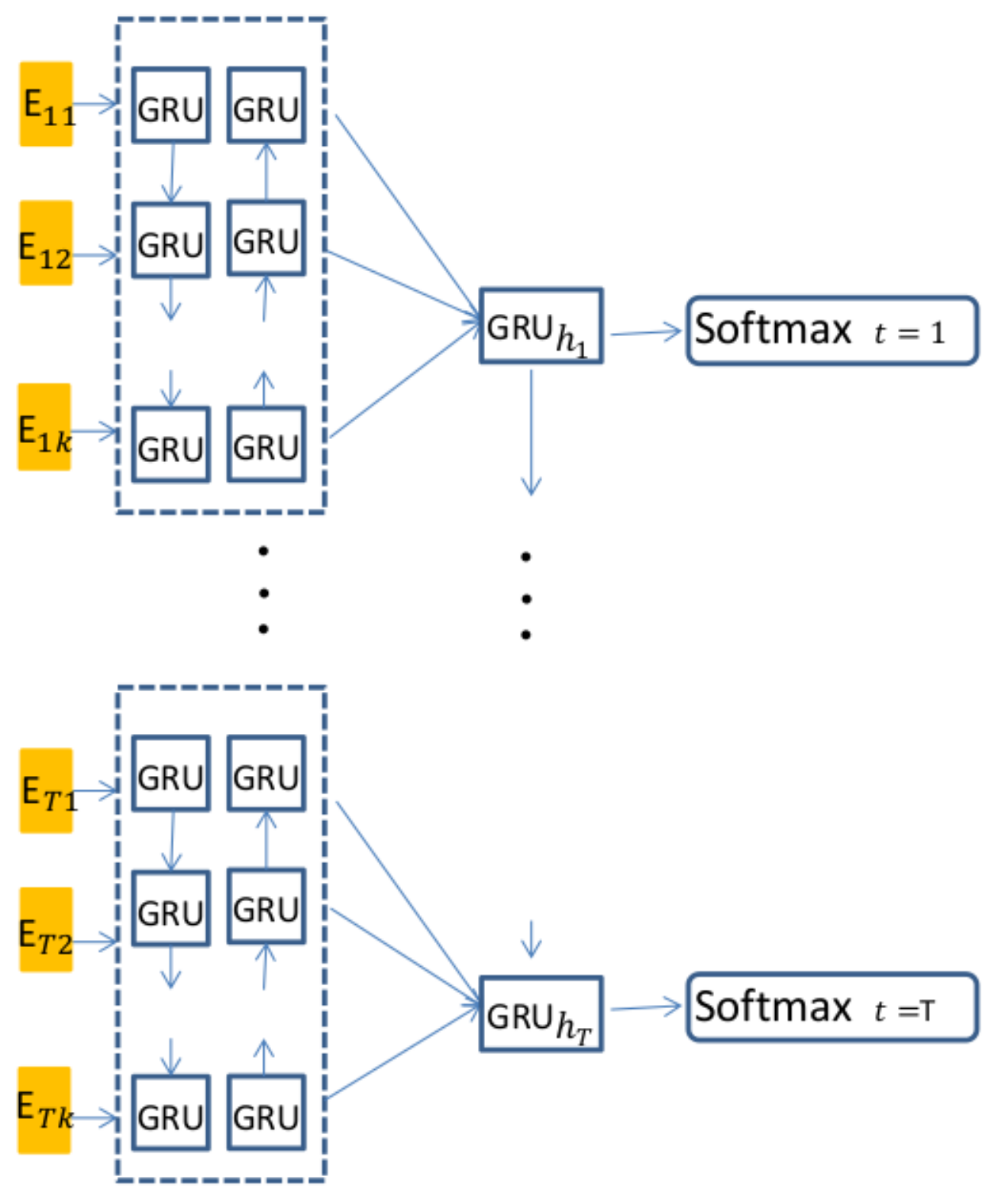}}
\subfloat[Contextual embeddings+GRU]{\includegraphics[scale=0.4]{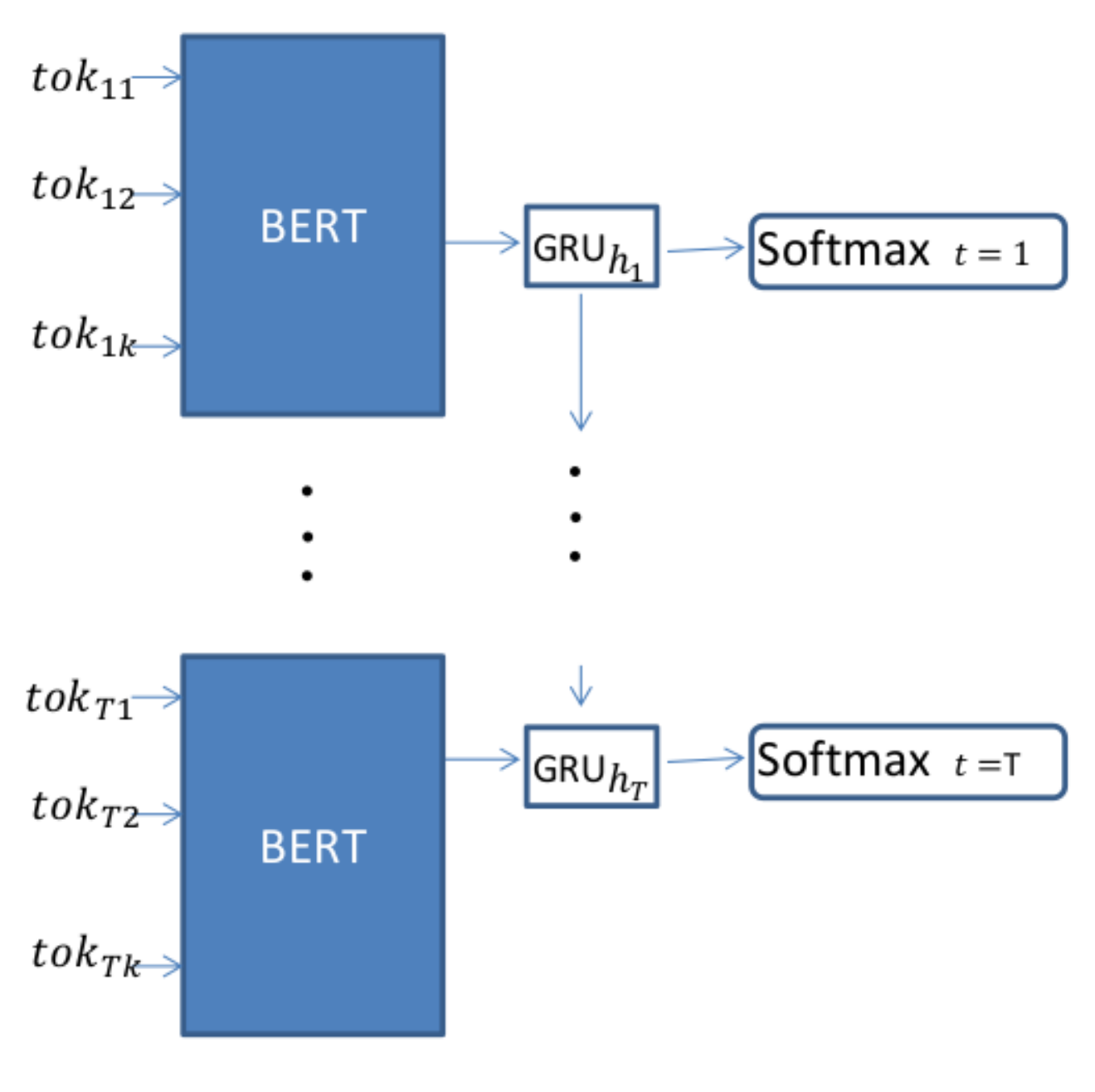}}
\subfloat[Transformer]{\includegraphics[scale=0.4]{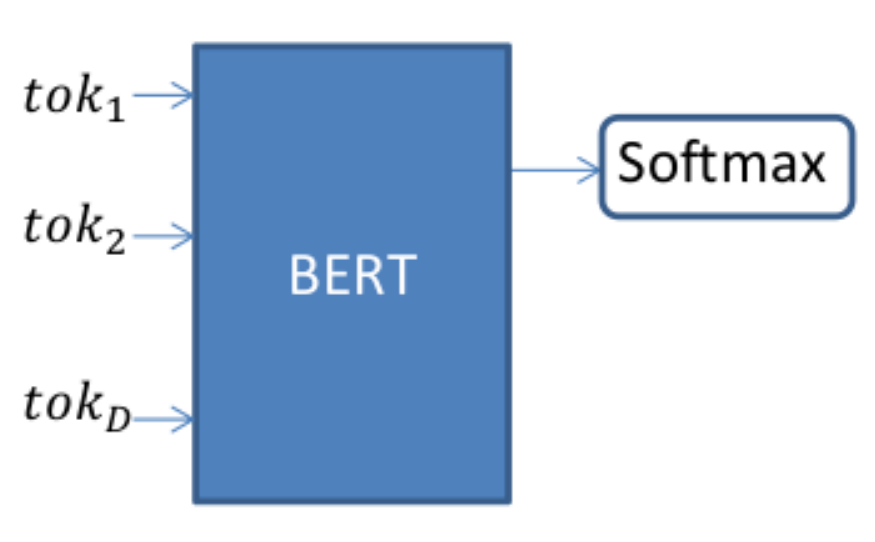}}

\caption{The proposed neural architectures for predicting the user satisfaction. $E_{Tk}$ in (a) is the embedding for the $k$th token of  the last turn ($t=T$). In (b) $tok_{T1}=\mbox{[CLS]}$ and  $tok_{Tk}=\mbox{[SEP]}$ after applying the WordPiece tokenization to the last turn ($t=T$). In (c) $tok_1,...,tok_D$ are the tokens of the dialogue after applying the WordPiece tokenization, in which $tok_1=\mbox{[CLS]}$ and the token [SEP] marks turn separation. $tok_{D-1}$ is the last token of the last utterance in the dialogue and $tok_D=\mbox{[SEP]}$.}
\label{f:bigru}
\end{figure}

\paragraph{\textbf{(i) Hierarchical GRUs:}}  Figure~\ref{f:bigru}(a) shows the network. It has a Bidirectional GRU layer (BiGRU) at the lower level that returns the turn representation $h_{tk}$ (Eq. \ref{eqn:htk}). 

\begin{align}
    \vec{h_{tk}}&=\mbox{GRU}(E_{tk},\vec{h_{tk-1}})\label{eq:htk_lr}\\
    \cev{h_{tk}}&=\mbox{GRU}(E_{tk},\cev{h_{tk-1}})\label{eqn:htk_rl}\\
    h_{tk}&=[\vec{h_{tk-1}},\cev{h_{tk}}]\label{eqn:htk}
\end{align}
Attention is used to weight relevant units in the turn hidden representation (Eq. \ref{eq:ha}) .
\begin{align}
h^a_{tk}&=\mbox{tanh}(W_{tk} \cdot h_{tk})
\\
\alpha_{tk}&=\mbox{softmax}(\gamma(W^a \cdot h^a_{tk}))\\
h^a_t&=\alpha \cdot h_{tk}\label{eq:ha}
\end{align}
A GRU layer is then used to process dialogues as a sequence of turns.
\begin{equation}
    h_t=\mbox{GRU}(h^a_{t},h^a_{t-1})
    \label{eq:GRU}
\end{equation}
The last layer is a Softmax that predicts the IQ score:
\begin{equation}
P(Y_t=c|h_t,W,b) = \frac{e^{(W_{c}h_t+b_c)}}{\sum_{c\prime} e^{(W_{c\prime}h_t+b_c\prime)}}
 \label{eq:softmax}
\end{equation}
where $c\in [1,..,\mathbf{C}]$ is the index of the output neuron representing one class ($\mathbf{C}=5$ for IQ scores).
The predicted score is the most probable class:
\begin{equation}
 \hat{y_t}=\mbox{argmax}_c(P(Y_t=c|h_t,W,b))
\end{equation}

\paragraph{\textbf{(ii) Contextual embeddings + GRU:}} We investigate the use of transformers as turn representations thus we propose a BERT-like transformer, which inputs are the tokens of the turn. The turn representations then feed a GRU layer (Eq. \ref{eq:GRU}).  The output of the GRU then feeds a Softmax layer (Eq. \ref{eq:softmax}) for predicting the score.

\paragraph{\textbf{(iii) Transformer:}} This network is depicted in Figure~\ref{f:bigru}(c). It consists in a transformer that takes as input the tokens {$tok_1,...,tok_D$} of the previous and current utterances. Then the output [CLS] of the transformer feeds a Softmax layer (Eq. \ref{eq:softmax}) for predicting the score of the current utterance. %In the Figure $tok_D$ is the last token of the last utterance of the dialogue.

We evaluate the prediction $y_t$ at each system turn $t$. The back-propagation optimisation is done by minimising the cross entropy loss function~\cite{tieleman2012lecture} through stochastic gradient descent.
%and BERT and DistilBert  bilingual embeddings.
%For treating French we used BERT and DistilBERT. For treating English we also explored extra-large transformers because bi-lingual pretrained embeddings were not yer available.
%A flat model that uses only DistilBert and an dialogue context of up to 512 tokens was also proposed for the prediction of the NPS score.

%We explore extra large transformer (transformer-xl) since dialogues contains more than 50 dialogue turns in average and the limit of 512 tokens of BERT and DistilBERT were not enough to cover the dialogue context.

\section{The Reward Function in a POMDP Dialogue System}
\label{pomdp}
Reinforcement Learning has long been used for learning dialogue strategies~\cite{levin2000stochastic,cuayahuitl2009hierarchical,rieser2011reinforcement}.  Dialogue is then formulated as an optimisation problem in which the final goal is to maximise the accumulated reward at long run~\cite{rieser2011reinforcement}.  The \textit{reward} is received from the environment (i.e. the user). The optimal {\it policy} is a function that takes as argument the current state $s$ and returns the optimal action $a$. 
POMDPs is an outstanding way of modelling dialogue~\cite{young2013pomdp,Casanueva2017} in which the state is uncertain, namely the belief-state.

\begin{figure}[ht]
	\centerline{\includegraphics[scale=0.5, keepaspectratio]{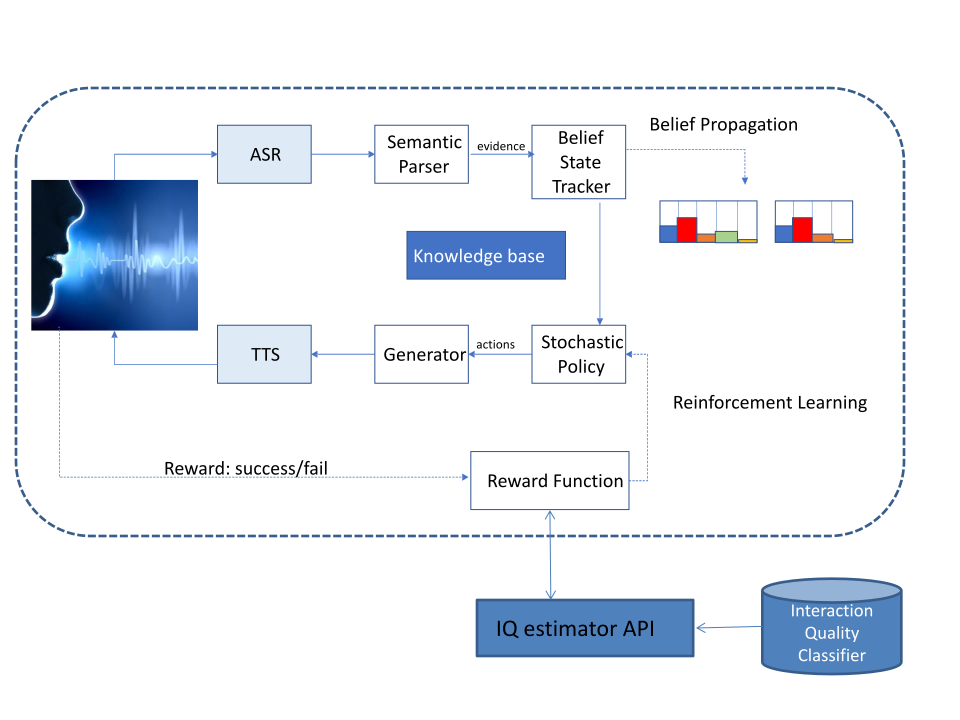}}
	\caption{\small\noindent A POMDP spoken dialogue system with the interaction quality estimator.}
	\label{achictecture}
\end{figure}

The basic elements of a POMDP dialogue system are shown in Figure~\ref{achictecture}. The words recognised by the speech recognition are converted to an abstract representation (the user dialogue acts).  These user dialogue acts are then processed by a belief tracker which maintains
a {\it dialogue state} $s$.  This is typically a set of variables denoting the {\it slots} that the system must fill-in to complete the user's goal. For example, in the Let's Go bus-scheduled information system, the slots might 
be \attr{origin} for the bus departure place and \attr{time} for the departure time. The state $s$ might record the current value and confidence level of each slot. From the state, a {\it belief state} $b$ (usually just a sub-set of the state vector) is extracted and an action $a$ is decided based on a dialogue {\it policy}.  The set of possible actions will include requesting new slot values, confirming already filled slot values and accessing the application for information. Once the appropriate action is determined, it is converted to a textual message and then rendered by a speech synthesiser.

The reward function most commonly adopted for task-oriented dialogues penalises every dialogue turn with $-1$ and sums a reward of $+20$ at the end of the dialogue whenever the system provided the right information to the user or $0$ otherwise (Eq.~\ref{eq:tsrwd})~\cite{gasic2013gaussian}.
\begin{equation}
    R_{TS}=T\cdot(-1)+\mathbbm{1}_{TS}\cdot 20
    \label{eq:tsrwd}
\end{equation}

In this work the reward estimator is based on the IQ as defined in~\cite{ultes-2019-improving}.

\begin{equation}
    R_{IQ}=T\cdot(-1)+(iq-1)\cdot 5
    \label{eq:iqrwd}
\end{equation}
Where $R_{IQ}$ describes the final reward, $iq$ is the IQ value predicted by the classifier (Section~\ref{estimators}), which is a number from 1 to 5, where 1 represents poor quality and 5 good quality.

We used PyDial~\cite{ultes2017pydial}, the publicly available POMDP dialogue framework and we implemented an application programming interface (API) that returns the IQ estimation predicted by the neural models presented in Section~\ref{estimators}.
Usually RL systems first learn the policy on a simulated user (User Simulator) until an optimal performance is reached, then they are ready to be tested by humans.
%RL dialogue systems require to simulate the user (User Simulation) during the learning because at early stages of training no human will stand chatting with it.
%which is a less subjective score  with less variance than self-reported scores~\cite{schmitt2011modeling}.
\section{Experiments}
\label{exps}
In this section we introduce the corpus as well as describe the experiments and the evaluation metrics.

\subsection{The Dataset}
The LEGO corpus collects \textit{spoken} dialogues between users and the Let's Go dialogue system~\cite{raux2005let}, which provides bus schedule information to the Pittsburgh population during off-peak times. 400 dialogues in the corpus have been manually annotated with the IQ score~\cite{schmitt-etal-2012-parameterized}. Since conversations in LEGO are system-initiative they have a lot of system interactions such as misunderstandings, confirmations and repetitions, producing quite long dialogues (i.e. hundreds of turns).

\begin{table}[htbp]
   \caption{The LEGO corpus with IQ annotations. Its complexity is measured by the maximum dialogue length (number of turns per dialogue), the maximum turn length (number of tokens per turn) and the maximum number of tokens per dialogue.}
 \label{t:corp}
\centering
  \begin{tabular}{c|c|c|c}

  N. Dialogues&Dialogue length
  &Max.turn length & Max. toks p/dial.\\[-1ex]
  &\scriptsize{(max/mean/median)}&\scriptsize{(max/mean/median)}&\scriptsize{(max/mean/median)}\\\hline
  400 & $200/65/53$& $76/26/24$& $6590/1444/1089$\\
  %15200?
  %\scriptsize{DATCHA}&\scriptsize{Technical support}&\scriptsize{Fr}&\scriptsize{NPS/dialogue-level}&\scriptsize{H-H}& \scriptsize{$183$}& \scriptsize{$300$}& \scriptsize{$54900$}\\
 \end{tabular}

 \end{table}

The complexity of the corpus is presented in Table~\ref{t:corp}, containing long dialogues of up to 200 turns with up to 76 tokens per turn.

\subsection{Hyperparameters and Training}
Dropout was used on BiGRUs networks (a dropout rate of $0.5$) to prevent co-adaptation of hidden units by randomly dropping out a proportion of the hidden units during forward propagation~\cite{hinton2012improving}. The models were implemented in PyTorch~\cite{paszke2017automatic}. To initialise the hierarchical models (BiGRUs) we use FastText embeddings~\cite{bojanowski2017enriching} with a dimension $d=300$.
For contextual embeddings we used the Transformers HuggingFace library~\cite{Wolf2019HuggingFacesTS} with the pre-trained embeddings and Wordpiece tokenisation. We used Adam optimiser~\cite{kingma2014adam}. All the models were trained on GPU machines with maximum of 32GB per GPU. We do not use TPUs.
%We use the AI high performance computing Jean Zay.

\subsection{User Satisfaction Estimators}
\label{ss:us}
%In this section we present the performance of the networks for estimating the scores IQ on the LEGO.

%\subsubsection{Estimation of the Interaction Quality}
%\label{sss:iq}

We compared our networks with the networks presented in~\cite{ultes-2019-improving} for predicting the IQ with the following evaluation metrics: the unweighted average recall (UAR), which is the arithmetic average of all class-wise recalls, as well as a linearly weighted version of Cohen's $\kappa$ and Spearman's $\rho$. The experiments were conducted in a 10-fold cross-validation, assuring that the same dialogue did not slip into different folds (i.e. dialogue-wise cross validation).  We studied the length of the \textbf{dialogue context}: the turn for which we are predicting the score and the previous turns. We vary the context length and found an optimal context length of up to 100 turns per dialogue for BiGRUs.

Table~\ref{t:iq} shows that our BiGRUs network trained on word embeddings outperforms the state-of-the-art (BiLSTM+att) networks in all performance measures, obtaining an absolute improvement of $+1$ for UAR, $+9$ for $\kappa$ and $+2$ for $\rho$. It is worth noting that the state-of-the-art (BiLSTM+att) networks were trained on fine-tuned \textit{turn features}. 

These results are encouraging and suggest that distributed representations impacts positively the performance of satisfaction estimators in hierarchical networks. We would like to study in the next section whether these models can be used to predict task success in dialogue systems.

%In LEGO dataset the maximum dialogue length was 200, due mainly to repetitions and confirmation of the Let's Go system.

\begin{table}[htbp]
 \caption{Performance of the proposed models. The BiGRUs with FastText embeddings outperform all the networks trained on fine-tuned \textit{turn features}}.
 %The star denotes statistical significance with the Wilcoxon test ($p < 0.005$)
 \label{t:iq}
\centering
  \begin{tabular}{l | lllll  }
 \hline
 \multicolumn{4}{c}{Predicting IQ}\\\hline
 Model&UAR&$\kappa$&$\rho$\\\hline
SVM\_feats~\cite{ultes-2019-improving} &$44\%$&$53\%$&$69\%$\\
BiLSTM+att\_feats~\cite{ultes-2019-improving} &$54\%$&$65\%$&$81\%$\\
BiGRUs&$\mathbf{\mathbf{55}}\%$&$\mathbf{74}\%$&$\mathbf{83}\%$\\
DBert+GRU&$35\%$&$57\%$&$42\%$\\
Trans-XL(ctxt$\approx$1K)&$47\%$&$59\%$&$67\%$\\
Trans-XL(ctxt$\approx$2K)&$44\%$&$58\%$&$67\%$\\
 
 \hline
 \end{tabular}

 \end{table}

BERT-based Transformers do not perform well for this task on this dataset when using them to get the turn representations, namely setting (ii) in Section~\ref{estimators}. This can be explained by the large number of turns dialogues have (i.e., up to $200$ turns), the large number of parameters a transformer needs and the quite short annotated dataset ($\approx$ 400 dialogues). We first tried (BERT+GRU) and due to its large memory requirement we could only learn weights of up to $7$ dialogue turns per dialogue in a cluster of 32GB-GPU machines (all the other turns representations were frozen). However, with DistilBERT we could treat up to $15$ turns per dialogue while maintaining the same performance. Fortunately, we could process larger contexts with Transformers-XL (setting (iii) in Section~\ref{estimators}), reaching an optimal performance with $\approx 1K$ dialogue tokens. The results of Transformers-XL are comparable with the SVM baseline trained on fine-tuned features, yielding a better UAR ($+3$) and $\kappa$ ($+6$) as well as a slightly lower $\rho$ ($-2$). Having larger contexts $\approx 2K$  do not seem to impact significantly their performance.
These results suggest that in 32GB-GPU nodes the large context length (i.e. up to $6.5$K tokens per dialogue) is affecting transformers performance as they will require a prohibited usage of GPU memory to process the whole context.

\subsection{The Impact on the Reward Function}

We evaluated the impact of the IQ estimators presented in Section~\ref{ss:us} on the reward function for the Let's Go (LetsGo) domain by using PyDial~\cite{ultes2017pydial}.

 As shown in Table~\ref{t:domains}, the Let's Go dialogue system provides information about bus time-schedule according to the constraints: \attr{origin}, \attr{destination}, \attr{time} and  \attr{route}, which corresponds to LetsGo(4). LestGo(6) also considers \attr{origin neighbourhood} and \attr{destination neighbourhood}. It is important to note that the Let's Go domain is far more complex in terms of the number of database items than other domains available in PyDial.
 %Cambridge Restaurants, CR(3), provides information about restaurants according to three constraints: \attr{food}, \attr{area} and \attr{price-range}. Besides the number of constraints, we can see that the Let's Go domain is far more complex than CR in terms of the number of database items.
 
 \begin{table}[ht]
      \caption{The Let's Go domain was used to compute the reward function. Note that it is far more complex in terms of the database items than other domains available in PyDial such as Cambridge Restaurants .}
     \label{t:domains}
     \centering
     \begin{tabular}{llll}\hline
     \textit{Domain}&\textit{\#constraints}&\textit{\#DB items}\\\hline
    LetsGo(4)&$4$&$100000$  \\
     LetsGo(6)&$6$&$100000$ \\
    CamRestaurants(3)&$3$&
     $110$\\\hline
     \end{tabular}

 \end{table}

 The experiments run on simulated dialogues as in~\cite{ultes-2019-improving,Casanueva2017}.
 We implemented a template-based generator for the user and the system utterances for the Let's Go domain because our models rely on textual inputs (i.e. distributed representations).  %The user simulator in PyDial usually works merely with the user dialogue-act input (i.e. semantic-level user simulation).
 We compared our models in an environment without noise because unlike~\cite{Casanueva2017} and~\cite{ultes-2019-improving} the simulator in this work runs at the surface-level and not at the semantic-level and the noise used for User-Simulation in PyDial alters the semantic-dialogue acts regardless the surface form. In addition, we would like to apply these methods to chatbots, thus simulating ASR noise would not be appropriate and studying a more appropriated noise is out of the scope of this paper.
 
 \begin{table}[htbp]
  \caption{Task success rate of the simulated experiments for the Let's Go domain over three runs with distinct seeds. Each value is computed after $1000$ training dialogues/$100$ evaluation.}.
 %(known) and the Cambridge Restaurant (unknown) domains. Each value is computed after $1000$ training dialogues/$100$ evaluation.}. 

 \label{t:tsuc}
\centering
  \begin{tabular}{l| lll }
 \hline
 %c |
 %&\scriptsize{Type}
 Domain&Reward&Task Success Rate($\uparrow$)& Average Turns($\downarrow$)\\\hline
 %&\multirow{4}{*}{\scriptsize{Known}}
 LetsGo(4)&$R_{TS}$&$\mathbf{99\%\pm1.4}$&$6.5\pm0.96$\\ %[-1ex]
 &$R_{IQ}$&$\mathbf{99\%\pm2.1}$&$\mathbf{6.3\pm0.8}$\\\hline
 LetsGo(6)&$R_{TS}$&$81\%\pm7.78$&$11\pm1.09$\\ %[-1ex]
 &$R_{IQ}$&$\mathbf{97\%\pm3.98}$&$\mathbf{8.9\pm1.26}$\\\hline
 % \scriptsize{CR (3)}&\scriptsize{$R_{TS}$}&\scriptsize{$\mathbf{97\%\pm0.99}$}\\ [-1ex]
 %&\scriptsize{$R_{IQ}$}&\scriptsize{$44\%\pm25.74$}\\ \hline
 \end{tabular}
%\multirow{2}{*}{\scriptsize{Unknown}}&
 \end{table} 
 
 We used a policy model based on the GP-SARSA algorithm~\cite{gasic2013gaussian}, which is a sample efficient Gaussian process approximation to the value function. We used the focus tracker~\cite{henderson2014second} for belief tracking. The policy decides on summary actions of the dialogue state tracker which are based on dialogue acts (e.g., \textit{request}, \textit{inform} or \textit{confirm}). \textit{The task success rate} was the metric used to measure the dialogue performance~\cite{Casanueva2017}.
 
We observe in Table~\ref{t:tsuc} that the reward computed with $R_{IQ}$ outperforms the classical $R_{TS}$ reward when having more constraints, namely LetsGO (6). Moreover, dialogues rewarded by $R_{IQ}$ tend to be significantly shorter. Although there is not significant distinction between $R_{TS}$ and $R_{IQ}$ in terms of the task success for LetsGO (4), dialogues are slightly short with $R_{IQ}$.  
We also conducted preliminary experiments on domain transfer by evaluating $R_{IQ}$ on the Cambridge Restaurants domain, obtaining a success rate of $44\pm25.7$, compared to $99\pm1.83$ for $R_{TS}$. Unsurprisingly, $R_{TS}$ and feature-based $R_IQ$~\cite{ultes-2019-improving} are more robust to unknown domains than embedding-based $R_{IQ}$ because both task-success and quality features are domain-agnostic.
%that for the same domain the estimator was trained on (Let's Go),
%We also observed that $R_{TS}$ is more robust to unknown domains. According to~\cite{ultes-2019-improving}, $R_IQ$ based on models trained on \textit{turn features} is also robust to unknown domains. However, turn features are not always available when using ASR services and are not directly applicable to chatbots. Typically, we only have access to the recognised or typed user's words.

\section{Discussion and Conclusions}
\label{disc}
We presented in this paper deep learning architectures for the estimation of the user satisfaction that exploits semantically rich distributed vectors. We found that distributed representations greatly improve the performance of hierarchical networks (BiGRUs) in all the measures (UAR, $\kappa$ and $\rho$) for estimating turn-level scores. 

Moreover, we studied different models by using Transformers-based representations. The results of our experiments suggest that in 32GB-GPU machines the large context length (i.e. up to $6.5$K tokens per dialogue) is affecting transformers performance as they will require a prohibited usage of GPU memory to treat the whole context.  These results are in line with findings and claims presented in~\cite{henderson2019convert} regarding the usage of Transformers for real-world conversational applications. Using Transformers with a limited-token context ($\approx 1K$) perform better in terms of UAR and $\kappa$ than the SVM baseline presented in~\cite{ultes-etal-2015-quality}.  Hierarchical networks (BiGRUs) not only performed better but also allowed us to study the impact of context by varying the dialogue length. We found that having a context of up to $100$ dialogue turns yields to an optimal performance. 

%So far adopting models trained on human dialogues with dialogue-level scores to compute the reward function seems unfeasible in a POMDP dialogue framework because of the complexity of the domain compared to information-seeking domains.%\footnote{\label{dialcompl}Dialogues for tasks that require a high dimensional state-action space.}.

%We have shown that BERT-based Transformers do not perform well for this task on dialogue corpora. In this work we studied two distinct dialogue datasets of different sizes in terms of the dialogue and turn lengths. As depicted in Table~\ref{t:corp}, in DATCHA the maximum dialogue length was $183$ turns, and the maximum turn-length was $300$ tokens. For LEGO the maximum dialogue length was $200$ turns and the maximum turn length $76$ tokens for a total of $54900$ and $15200$ tokens per train instance in DATCHA and LEGO respectively (Table~\ref{t:corp}). These will require a prohibited usage of GPU memory.

Furthermore, we studied the impact of applying hierarchical networks to compute the reward function $R_{IQ}$ in a POMDP dialogue system. We found that $R_{IQ}$ trained on distributed representations is more robust to a larger number of constraints than $R_{TS}$. However, either $R_{TS}$ or $R_{IQ}$ trained on turn features(\cite{ultes-2019-improving}) are still more robust to unknown domains because
by definition they are domain-agnostic.

We have shown that applying distributed-based models to compute the reward function can be a viable alternative for rewarding RL dialogue systems at runtime in the case ASR features are not available such as in chatbots. Therefore, one can devise rewarding dialogues at runtime when models can be trained on a human-human corpus with user-rated interactions.

\section{Future Work}
\label{future}
As future work we would like to investigate a new generation of Transformers (e.g. generative transformers GPT$-2$~\cite{radford2018improving}) for classification tasks applied to long dialogues. We also would like to further study domain transfer. We can also think in a reward function that combines satisfaction predictors, task-success and/or objective measures. In addition, we can study different kind of noise in the user simulation, besides ASR noise, for simulating users' inputs in chatbots (e.g. misspellings and typos).

%Modelling complex dialogues for technical-assistance, beyond information-seeking, within a RL dialogue framework is still an important research path.
%\footnoteref{dialcompl}
\section*{Acknowledgements}

We would like to thank all the members of the team DATA-AI/NADIA at Orange-Labs, especially  the former undergraduated students: Benjamin Lepers and Jean-Baptiste Duchene. We also thank Timothy Garwood and Stefan Ultes for the fruitful discussions. 

We earned unlimited access to the French HPC Jean Zay (IDRIS-CNRS)\footnote{\url{http://www.idris.fr/annonces/annonce-jean-zay.html}} with the project 10096 selected in the French contest \textbf{Grands Challenges IA 2019}. Therefore, all the experiments presented in this paper were run in Jean Zay HPC.

\bibliographystyle{IEEETran}
\bibliography{coling2020}

\end{document}